\documentclass[letterpaper, 10 pt, conference]{ieeeconf}  

\IEEEoverridecommandlockouts                               

\usepackage{amsmath} 
\usepackage{amssymb}  
\usepackage{textcomp}
\usepackage{gensymb}
\usepackage{graphicx}
\usepackage{caption}
\usepackage{subcaption}
\usepackage{cite}
\usepackage{color}
\usepackage{url}
\usepackage{bm}
\usepackage{array}
\usepackage{pgfplots}
\usepgfplotslibrary{groupplots}
\usepgfplotslibrary{external}
\usetikzlibrary{pgfplots.groupplots}
\usetikzlibrary{patterns} 
\usetikzlibrary{shapes,arrows,fit,calc,positioning,automata,backgrounds}
\usepackage{placeins}
\usepackage{multirow}
\usepackage[hidelinks]{hyperref}

\usepackage{times} 

\usepackage{amsbsy}  
\usepackage{algorithm}
\usepackage[noend]{algpseudocode}
\usepackage{color}			
\usepackage{xcolor}			
\usepackage{colortbl}		
\usepackage{rotating}

\usepackage{fancyhdr}
\fancypagestyle{withfooter}{
  
  \fancyfoot[C]{\footnotesize Accepted to the IEEE ICRA Workshop on Field Robotics 2024}
}

\definecolor{lightgray}{rgb}{0.9,0.9,0.9}

\title{\LARGE \bf
SVan: A Mobile Hub as a Field Robotics Development and Deployment Platform
}

\author{Alexander Moortgat-Pick, Anna Adamczyk, Daniel A Duecker, Sami Haddadin
\thanks{All authors are with Munich Institute of Robotics and Machine Intelligence (MIRMI), Technical University of Munich (TUM), Germany. 
Contact: {\tt\small author@tum.de}}
\thanks{The authors would like to thank the Dobeneck Technology Foundation for financial support as part of the SVAN: Synchronous Team-Robot Van project.
The authors acknowledge the financial support by the Federal Ministry of Education and Research of Germany (BMBF) in the programme of "Souverän. Digital. Vernetzt." Joint project 6G-life, project identification number 16KISK002.}
}

\begin{document}
\maketitle
\thispagestyle{withfooter}
\pagestyle{withfooter}


\begin{abstract}

As robotics becomes increasingly vital for environmental protection, there’s a growing need for effective deployment methods that match the pace of robotics innovation.
Current strategies often fall short, leaving a gap between the potential of robotics and their practical application in the field.
Addressing this challenge, we introduce a mobile hub concept designed to provide the necessary infrastructure and support for deploying a diverse, multi-domain robot team effectively.
This paper presents the development and insights into "SVAN" (Synchronous Team-Robot Van), a prototype of our mobile hub concept.
We delve into the mechanical construction and software setup of SVAN, offering a comprehensive overview of its capabilities and design considerations.
Further, we discuss the hardware specifications and share valuable lessons learned during the prototype's development and deployment.
In addition to this paper, an accepted video complements our exploration by depicting SVAN in its envisioned role as an environmental guardian, highlighting its potential in ecological monitoring and preservation.
Furthermore, our discussion is enriched by referencing a previously accepted paper detailing a novel methodology for continuous UAV mission cycling enabled by a mobile hub like SVAN.
These accompanying works underscore our contribution towards addressing the existing gaps in robot deployment strategies, presenting a scalable and efficient framework to overcome operational challenges in environmental robotics.
\end{abstract}

\graphicspath{{figs/}} 


\section{Introduction}\label{sec:intro}
\subsection{Motivation}\label{sec:motivation}

\begin{figure}
    \centering
    \includegraphics[width=0.5\textwidth, clip]{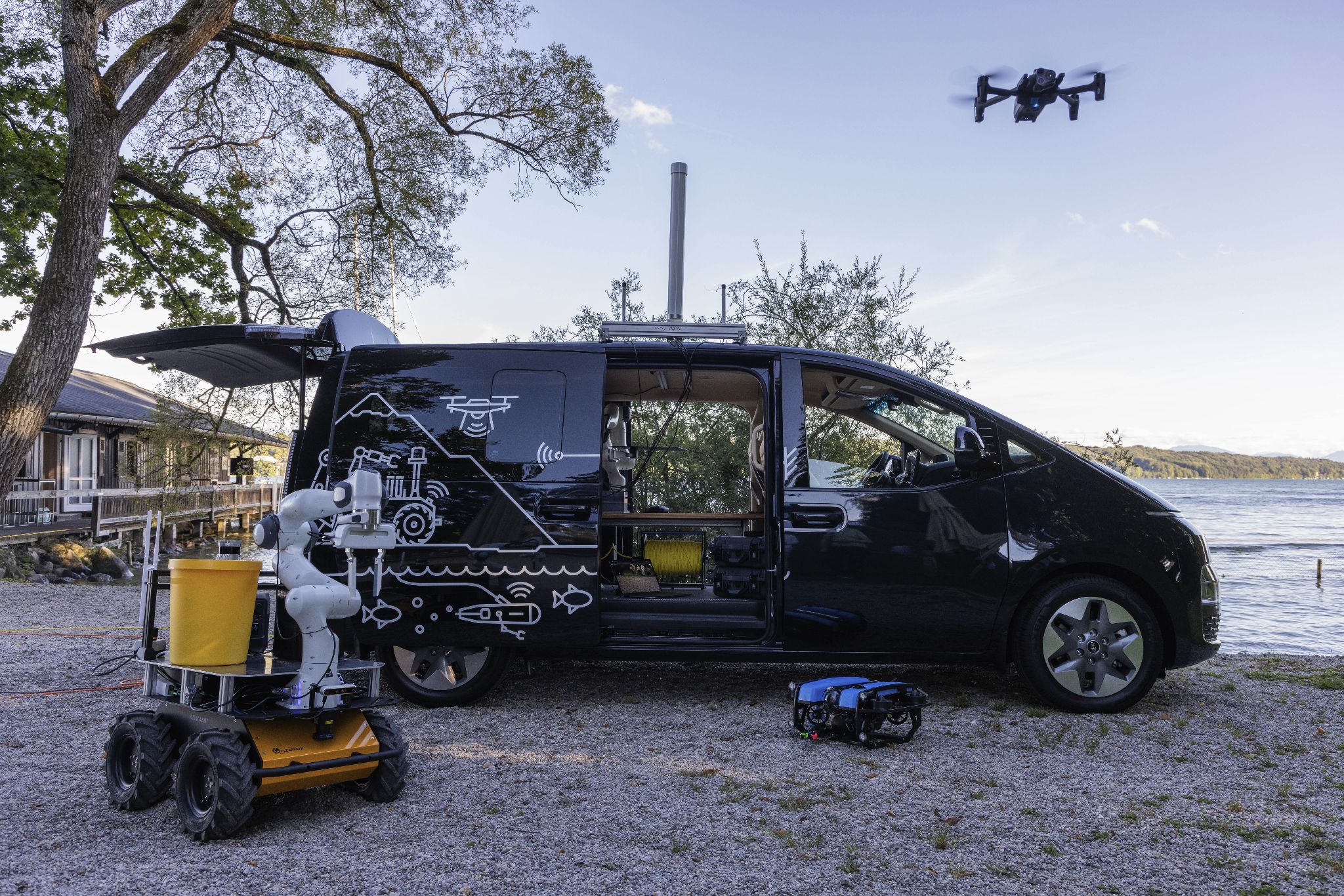}
    \caption[]{Our SVan demonstrator with supported robots in three domains. This mobile hub provides the entire infrastructure to perform multi-robot missions, including remote telepresence connections.}
    \label{fig:svan-and-all-robots}
\end{figure}

%
In the vanguard of environmental conservation, robotics has emerged as a pivotal ally, offering both innovative solutions and new challenges. The deployment of robots in outdoor environments is not only labor-intensive but also constrained by the perennial issue of limited budgets typical of environmental protection efforts. Our vision is to streamline this process, enabling a single individual to manage the deployment and upkeep of robots during missions efficiently. This approach not only optimizes resources but also aligns with the need for continual development and enhancement of our systems.

The logistic simplification of having an entire operational infrastructure on a single platform cannot be overstated. It represents the zenith of efficiency, significantly reducing the complexity of managing multi-domain robotic missions. These missions, which span various environments - from aerial to terrestrial and aquatic - require a versatile support system capable of addressing the unique demands of each domain.

To achieve this goal, we have identified our requirements as follows:
\begin{itemize}
\item Full-day operation: Ensuring that our platform can support extended missions without the need for frequent recharging or intervention.
\item Robot battery charging: Integral to maintaining operational continuity, allowing robots to recharge directly from the mobile hub.
\item Multi-domain robot support: A platform versatile enough to support and interact with robots operating in different environments and contexts.
\item Development platform: Providing on-the-go capabilities for system development, testing, and enhancement, facilitating immediate response to operational challenges.
\item AI enhancement of robots' capabilities (on the edge): Leveraging AI to augment the capabilities of robots in real-time, enhancing their effectiveness and adaptability.
\item Telepresence to and from the mobile hub: Enabling remote operation and interaction, expanding the operational scope beyond the immediate vicinity of the hub.
\item Data storage: Essential for documenting missions, collecting valuable data for analysis, and refining system performance.
\item Data synchronization with the cloud: Facilitating the secure and efficient transfer of data to and from cloud-based systems for enhanced access and collaboration.
\end{itemize}

\subsection{Related Work}\label{sec:related_work}

In the realm of field robotics, significant advancements have been made in enhancing the capabilities of robots for various applications. Notably, the integration of robots with supporting infrastructures, which facilitates extended operational capabilities and efficient data management, has seen limited exploration. A mobile data center designed to support search and rescue operations in shown in \cite{bedkowski2015support}. This innovation represents a crucial step towards integrating robotic systems with mobile infrastructure, enabling enhanced data processing and communication capabilities in critical missions.

A heterogeneous fleet of robots  - including a UAV, an AUV, and a USV - for the inspection of marine ecosystems was deployed by \cite{shkurti2012multi}. In this study, a fixed-wing UAV is utilized to collect continuous camera data, which is then transmitted over the Internet to scientists for analysis. The operation leverages a static infrastructure junction to relay data from remote robotic units back to researchers, illustrating the value of infrastructure in extending the reach and efficiency of robotic missions.

Similarly, \cite{ross2019collaboration} explored the collaborative use of UAV, AUV, and USV teams to investigate floating objects like ships and icebergs. Their concept, tested both in simulation and lab settings, underscores the potential of combining robotic fleets with infrastructure for complex investigative tasks. While their study underscores the significant potential of using collaborative teams of UAVs, AUVs, and USVs for investigating floating objects such as ships and icebergs, it also highlights a crucial limitation: the dependency on proper infrastructure.

The SeaClear project \cite{delea2020search} further exemplifies the synergy between robotics and infrastructure, employing a combination of ROVs, a USV, and an UAV to remove litter from ocean floors. Here, the USV plays a pivotal role as an infrastructure base, facilitating the coordination and operation of the other robots in the fleet.

Despite these advancements, the scientific literature reveals a relative scarcity of studies focused on the development and implementation of supporting infrastructure for robotic systems. 
It's understandable that in scenarios involving single robot applications \cite{bennetts2014robot} \cite{neumann2022remote}, or within research contexts staffed with specialized personnel \cite{neumann2019aerial} \cite{arain2021sniffing}, the necessity for robust supporting infrastructure might be overlooked. However, the practical experiences of deploying robotic teams, especially those comprising multiple types of robots, reveal a considerable logistical challenge. This challenge is particularly pronounced in the field of environmental monitoring, where missions are often designed to be of short duration yet require execution at various sites, with the base needing to be mobile, and with a high frequency of repetition throughout the year.

The logistical complexities associated with setting up and managing such missions underscore a significant gap in the current state of robotics. The need for a streamlined, efficient process for deploying robots in the field becomes increasingly critical.
It highlights the imperative for innovative solutions that address not only the technological but also the operational challenges of field robotics. This observation leads us to introduce our contribution: a novel approach designed to mitigate the logistical hurdles of field deployment, particularly for environmental monitoring missions that necessitate multiple, diverse types of robots and dynamic locations. An infrastructure base engineered for portability and on-site deployment flexibility.


\subsection{Contribution}\label{sec:contribution}

In this paper, we introduce the concept of a mobile hub for field robotics, realized through the development of SVAN - The Synchronous Team-Robot Van. This prototype has been actively utilized in our environmental telerobotics missions as an essential infrastructure base as well as more detailed research for autonomous UAV missions.
The aim of this paper is to detail the underlying concept of this mobile robot hub, the prototype SVAN and to share insights and lessons learned from its practical applications, highlighting its contribution to field robotics through the following real-world deployments.

SVAN's innovative application is highlighted in the ICRA 2024 Standalone Video, in the 'Imagining the Future' category, titled 'Robot Guardians: The Artificial Immune System of our Planet'. Here, SVAN is depicted as the cornerstone of field infrastructure, facilitating a future where robots act as guardians of Earth's ecosystems \cite{robotguardiansvideo2024}. Additionally, we showcased SVAN in a live demonstration, presenting our heterogeneous robot team's capabilities in conducting an outdoor simulated trash collection mission. This demonstration not only highlighted SVAN's hardware and software capabilities but also showcased its potential for enabling interactive human-robot interaction (HRI) sessions with visitors \cite{mirmi2024mobilehub}.

Additionally, our research of autonomous missions with a mobile hub delves into the autonomously cycling of UAVs through missions, utilizing the hubs robotic arm to ensure precise landings and continuous operations, even in challenging environments \cite{2024MoortgatPickUAVMissionCycling}. This exploration addresses key aspects of field robotics, demonstrating the mobile hub's critical role in enhancing the autonomy and resilience of UAV missions.

\section{Concept Mobile Robotics Hub}\label{sec:concept}

\begin{figure}[ht!]
    \centering
    \includegraphics[width=\columnwidth, clip]{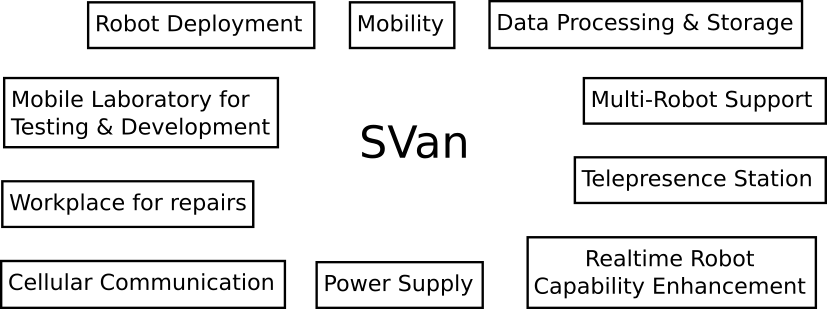}
    \caption[]{The SVAN concept includes the provision of a wide range of resources, infrastructure and capabilities, all by a single vehicle.}
    \label{fig:svan-concept}
\end{figure}

\begin{figure}[ht!]
    \centering
    \includegraphics[width=\columnwidth, clip]{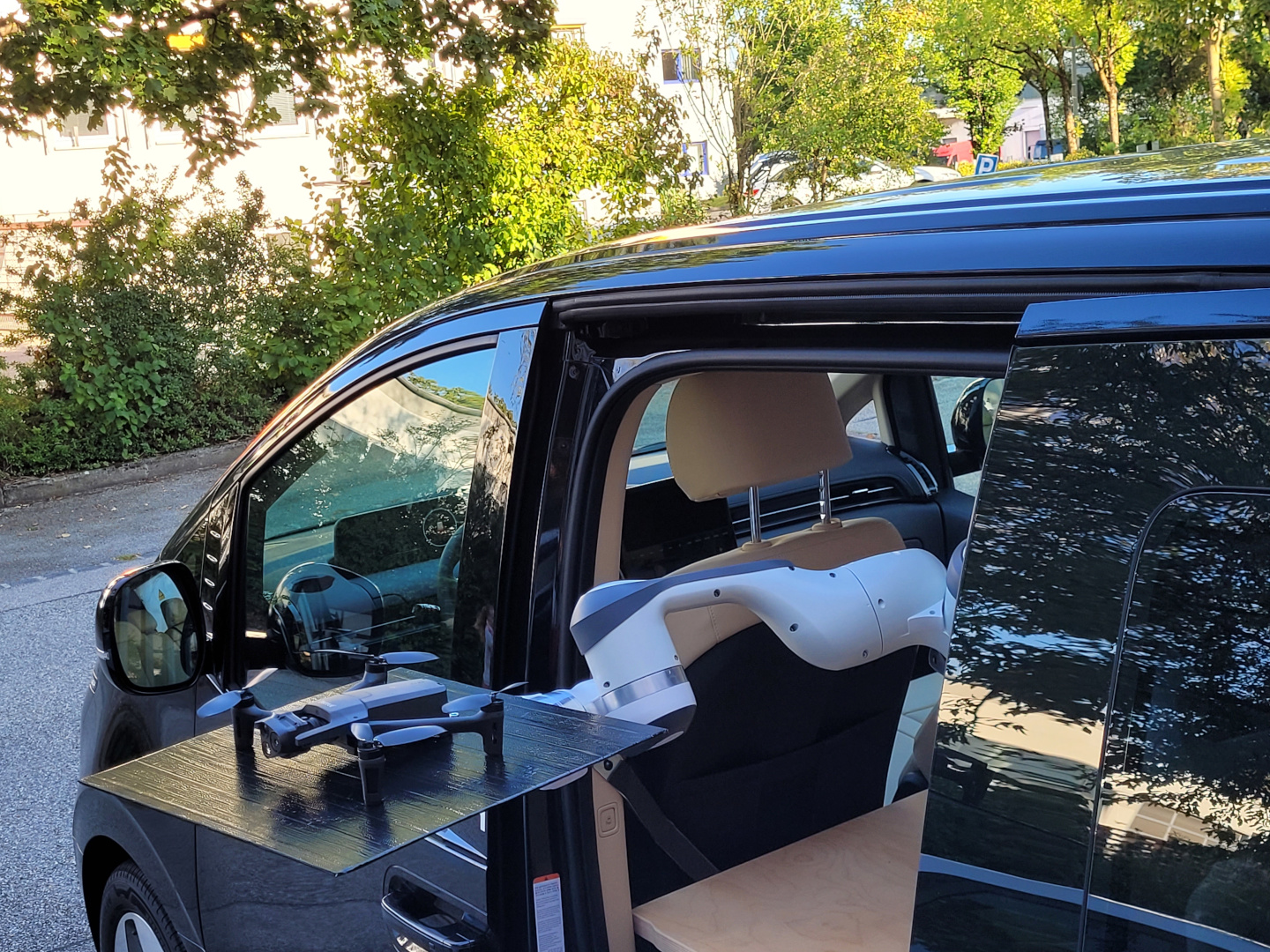}
    \caption[]{The hub concept supports missions by autonomously keeping UAVs in the air, cycling them between takeoff, mission, landing and maintenance.}
    \label{fig:svan-uav-launch}
\end{figure}

\begin{figure}[ht!]
    \centering
    \includegraphics[width=\columnwidth, clip]{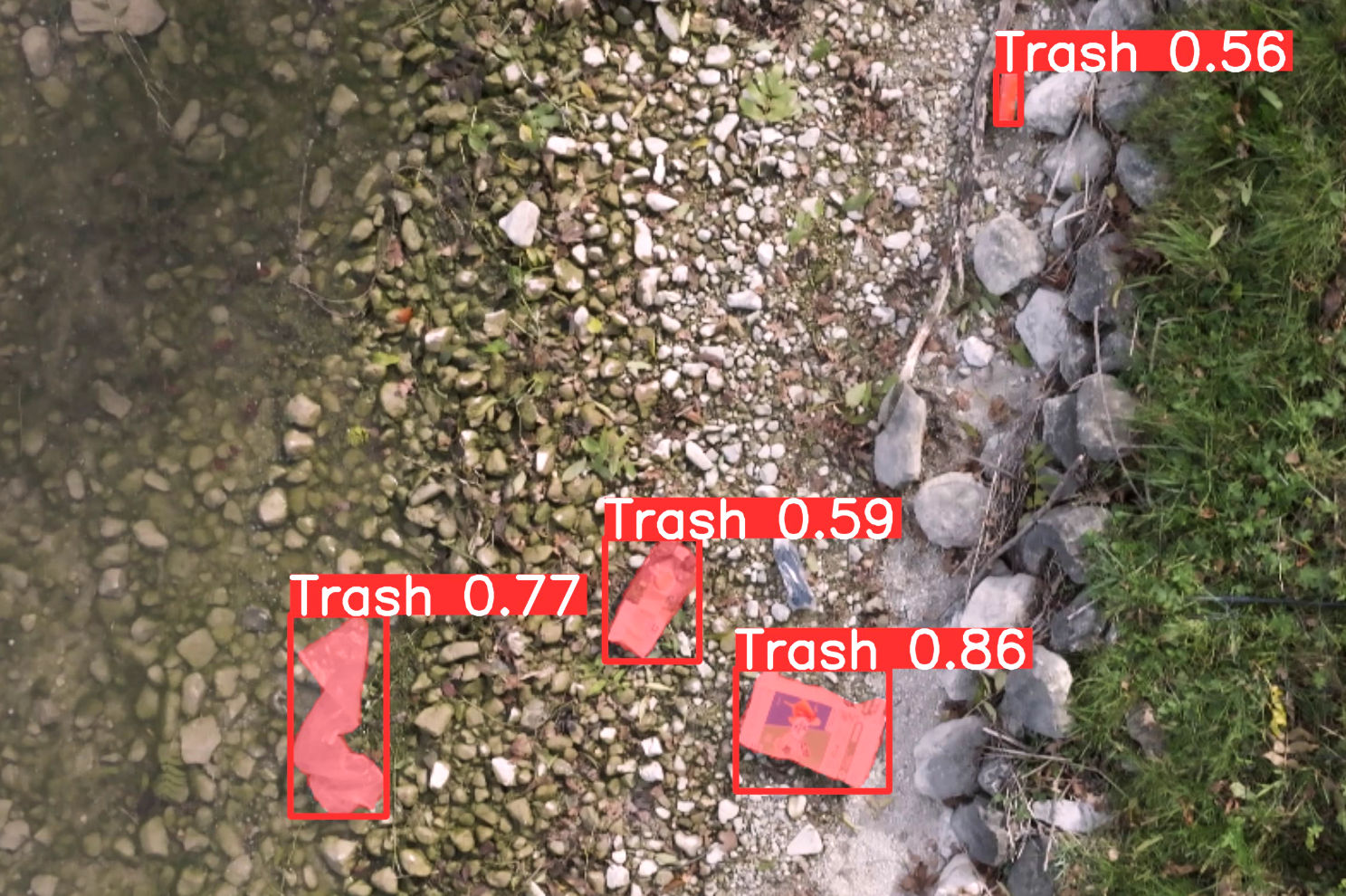}
    \caption[]{Trash detection in video feed as an example of edge AI provided by the hub as an extension of robot capabilities.}
    \label{fig:trash-detect-uav}
\end{figure}

\begin{figure*}[h!]
    \centering
    \includegraphics[width=\textwidth, clip]{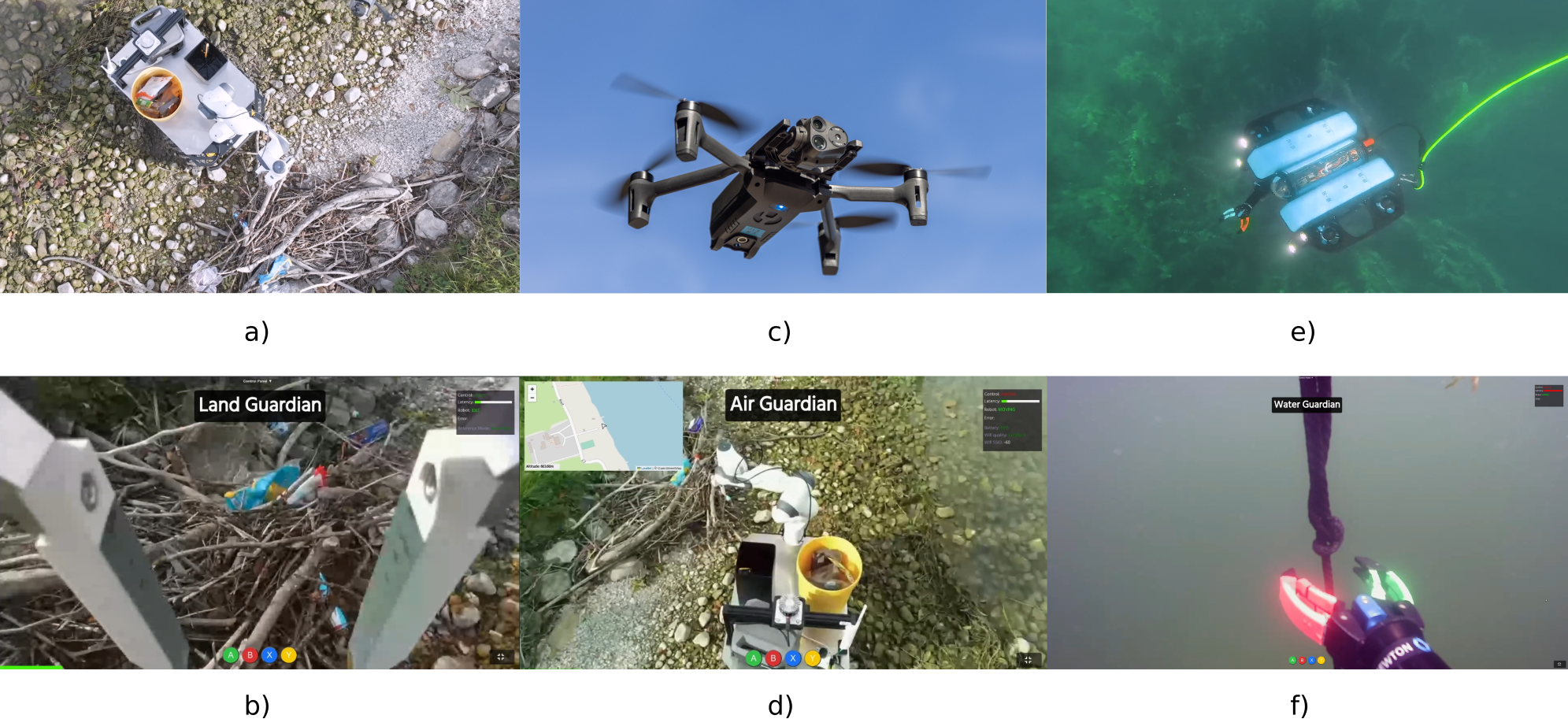}
    \caption[]{Multi-domain robots (top row) with corresponding user interfaces (bottom row). The left shows a land robot that is collecting trash. It can be used remotely via e.g. a smartphone. On the right an underwater robot can e.g. deploy buoys or also collect trash. The middle shows a UAV as supporting unit.}
    \label{fig:svan-robots}
\end{figure*}

Our innovative concept, as illustrated in Figure \ref{fig:svan-concept}, introduces a mobile robotics hub tailored for environmental robotics applications. This design is anchored in specific requirements outlined in Section \ref{sec:intro}, reflecting our targeted application scope. While acknowledging the specificity of our requirements, we posit that sharing our comprehensive framework can offer valuable insights and potential adaptation avenues for similar endeavors.

The SVAN hub is envisioned as a versatile "toolbox," equipped with an array of robots, machinery, computers, and materials, designed to facilitate a broad spectrum of outdoor robotics missions. Its functionality is bifurcated into two main categories: technical supply and human support.

\begin{figure}[ht!]
    \centering
    \includegraphics[width=\columnwidth, clip]{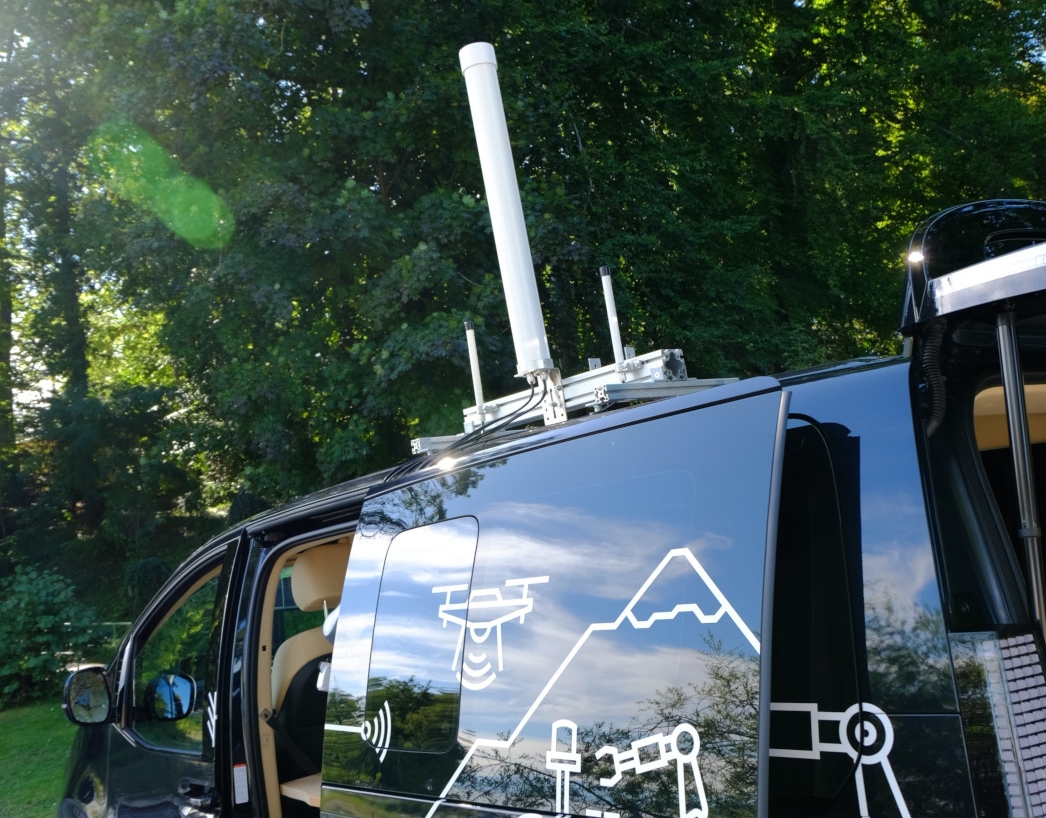}
    \caption[]{The wireless communication infrastructure consists of a 4x4 MIMO cellular and two WLAN antennas. Attached to a supporting frame, it can be easily attached on a vehicles roof with magnets.}
    \label{fig:svan-antenna}
\end{figure}

In terms of technical supply, the hub provides essential services such as power and communication resources, augmented data processing capabilities, and the potential for real-time enhancements to robot operations, exemplified by AI-powered trash detection algorithms. A pivotal feature is the integration of a high-quality cellular connection, facilitated by an enlarged antenna setup (Figure \ref{fig:svan-antenna}), ensuring reliable internet access even in secluded areas. Furthermore, the SVAN hub streamlines the deployment of robots, including an innovative approach for the autonomous launching of UAVs (Figure \ref{fig:svan-uav-launch}), as detailed in our previous work \cite{2024MoortgatPickUAVMissionCycling}.

Human support is equally critical, allowing a single operator to efficiently manage the SVAN hub and oversee mission execution, with a strong emphasis on leveraging autonomous deployment capabilities wherever feasible. The hub also serves as a pivotal resource during testing and development phases, offering a dedicated workspace for operators. This space is not only conducive to conducting minor repairs and calibrations but also supports data management and coding activities. Moreover, it can transform into a telepresence station, employing a robotic arm as a haptic interface, thus enabling intuitive control over various robotic systems across multiple domains.

This comprehensive approach not only addresses the practical necessities of environmental robotics missions but also fosters an integrated ecosystem conducive to innovation and operational efficiency.

\section{Implementation}\label{sec:implementation}

In detailing the implementation of the SVAN hub, it's essential to discuss both its mechanical construction and software setup, as both elements are pivotal to its functionality and efficiency.

\begin{figure}[ht!]
    \centering
    \includegraphics[width=\columnwidth, clip]{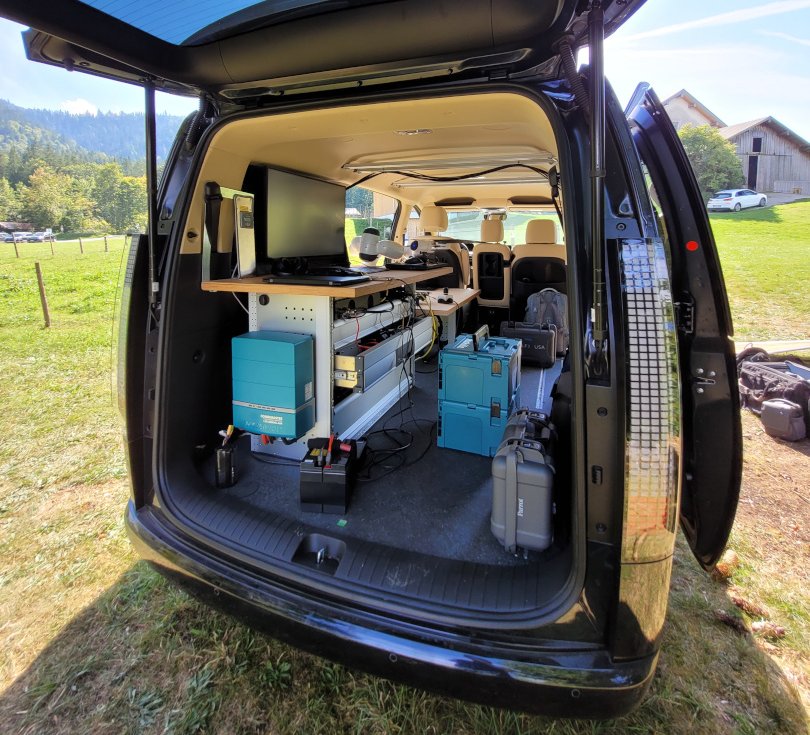}
    \caption[]{Interieur view of the mobile robot hub from the back.}
    \label{fig:svan-interieur-back}
\end{figure}

\begin{figure*}[ht!]
    \centering
    \includegraphics[width=0.9\textwidth, clip]{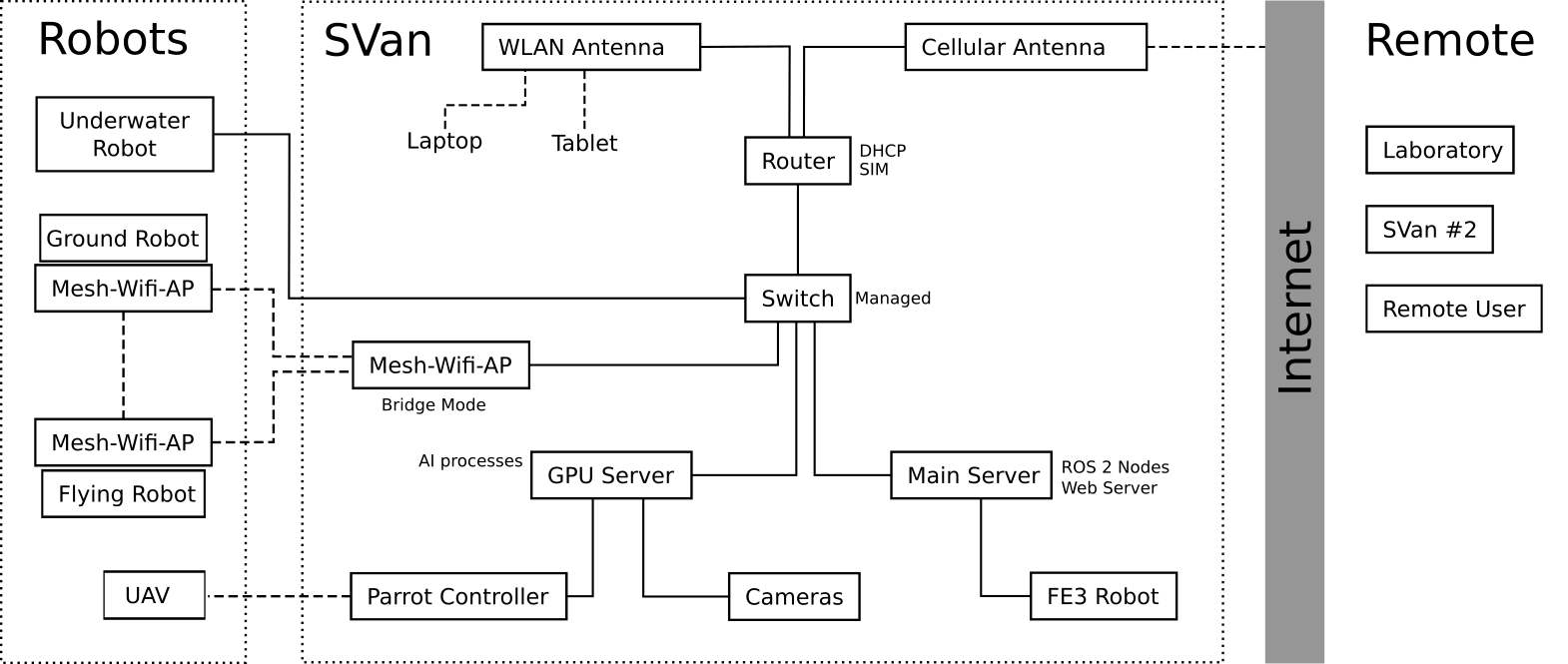}
    \caption[]{Overview of the networking setup. On top of the shown physical layer, a VPN can be used to create secure and simple networks with remote sites.}
    \label{fig:svan-networking}
\end{figure*}

\subsection{Mechanical Construction}
The transformation of a passenger vehicle into the SVAN hub required meticulous mechanical engineering. Given the inherent challenges of repurposing a vehicle not originally designed for such a task, the conversion process involved significant modifications. A specialized company was commissioned to retrofit the vehicle, which entailed removing the rear seating to make room for a new, robust flooring, the installation of two tables on the vehicle's left side, and an integrated power supply system. This comprehensive overhaul resulted in a stable and secure mobile platform, ideal for field operations. Figure \ref{fig:svan-interieur-back} showcases the vehicle's interior post-modification, highlighting the operator's workstation equipped with a large, vivid monitor connected to the hub's central computing unit.

\subsection{Software Setup}
On the software front, our primary focus has been on establishing a versatile and reliable networking infrastructure, illustrated in Figure \ref{fig:svan-networking}. At the core of the SVAN's networking capabilities is a central switch, facilitating seamless transitions between external antenna connections and a wired network interface. This flexibility is crucial for operations in varying environments, enabling the hub to connect to our lab's wired intranet during stationary phases and switch to antenna-based communications when mobile.

The networking architecture supports both wired and wireless connections within the hub, catering to the diverse needs of installed computers and portable devices such as smartphones and laptops. Communication protocols for the robots vary based on their specific requirements, ranging from direct wired connections for aquatic robots to proprietary Wi-Fi standards and a bespoke mesh Wi-Fi network for others.

A Virtual Private Network (VPN) layer overlays the physical network infrastructure, ensuring secure and efficient connectivity between the hub and remote locations. This setup facilitates the use of ROS 2 for intra-robot communication and integration with distant operational sites, enabling robust and flexible command and control capabilities across all deployed robotic assets.

Through this dual focus on mechanical robustness and advanced software infrastructure, the SVAN hub stands as a highly capable platform for environmental robotics, supporting a wide range of operational scenarios and technological requirements.

\subsection{Hardware Specification}

In providing an overview of the hardware specifications and capabilities of the SVAN hub, we detail both the robotic assets utilized for various missions and the foundational components of the hub itself, underscoring its operational versatility and technical sophistication.\\

The SVAN hub is equipped with an array of state-of-the-art robotics designed for diverse environmental tasks:
\begin{itemize}
\item BlueROV 2: A high-performance underwater vehicle from Blue Robotics, optimized for exploration and data collection in aquatic environments.
\item Parrot Anafi UAVs: Compact and versatile unmanned aerial vehicles, ideal for aerial surveillance and data gathering.
\item Clearpath Husky: A robust, all-terrain unmanned ground vehicle, serving as a versatile platform for outdoor navigation and task execution.
\item Franka Robotics Tactile Manipulators: Two advanced robotic arms, with one installed within the hub to facilitate UAV mission cycling and serve as a telepresence haptic interface. The second is mounted on the Husky, transforming it into a mobile manipulator capable of complex interaction and manipulation tasks.
\end{itemize}
These robots, each selected for their leading-edge capabilities in their respective domains as well as their ability to be integrated with our software system, are transportable within the SVAN hub, ensuring comprehensive mission support.\\

The hub itself has the following specifications:
\begin{itemize}
\item Vehicle Platform: The Hyundai Staria van, chosen for its ample interior space and a height of 1.99 meters, providing a compact yet spacious mobile base for the hub's operations.
\item Power Supply: The hub is powered by batteries boasting a 5.28 kWh capacity, supporting 5 to 10 hours of continuous operation, which is critical for extended field missions.
\item Operator Console: A 34-inch monitor combines high brightness and resolution, offering the operator a clear and detailed interface for monitoring and controlling missions.
\item Computational Resources: Two ruggedized computers, each with 16 CPU cores and a 35 W TDP, form the computational backbone of the hub. One computer is augmented with a GPU, bolstering its processing capabilities for demanding tasks such as AI computations and real-time data analysis.
\item Connectivity: A 4x4 MIMO antenna system, with a typical gain of 8 dBi, connected to a high-performance router, facilitates cellular connections. This setup is capable of achieving downlink speeds up to 500 Mbit/s and uplink speeds up to 200 Mbit/s. Operational tests in the lakes region of Bavaria, Germany, have demonstrated typical bandwidths of approximately 80 Mbit/s down and 20 Mbit/s up, surpassing the requirements for effective telepresence applications.
\end{itemize}

Through this harmonious integration of advanced robotics, robust vehicle modifications, and high-performance computational and communication systems, the SVAN hub emerges as a highly capable platform. It is designed to support and enhance environmental robotics missions, offering unparalleled flexibility and efficacy in field operations.

\section{Lessons Learned}\label{sec:lessons}
\subsection{Mechanical Construction}\label{sec:mechanical_construction}
\begin{figure}
    \centering
    \includegraphics[width=\columnwidth, clip]{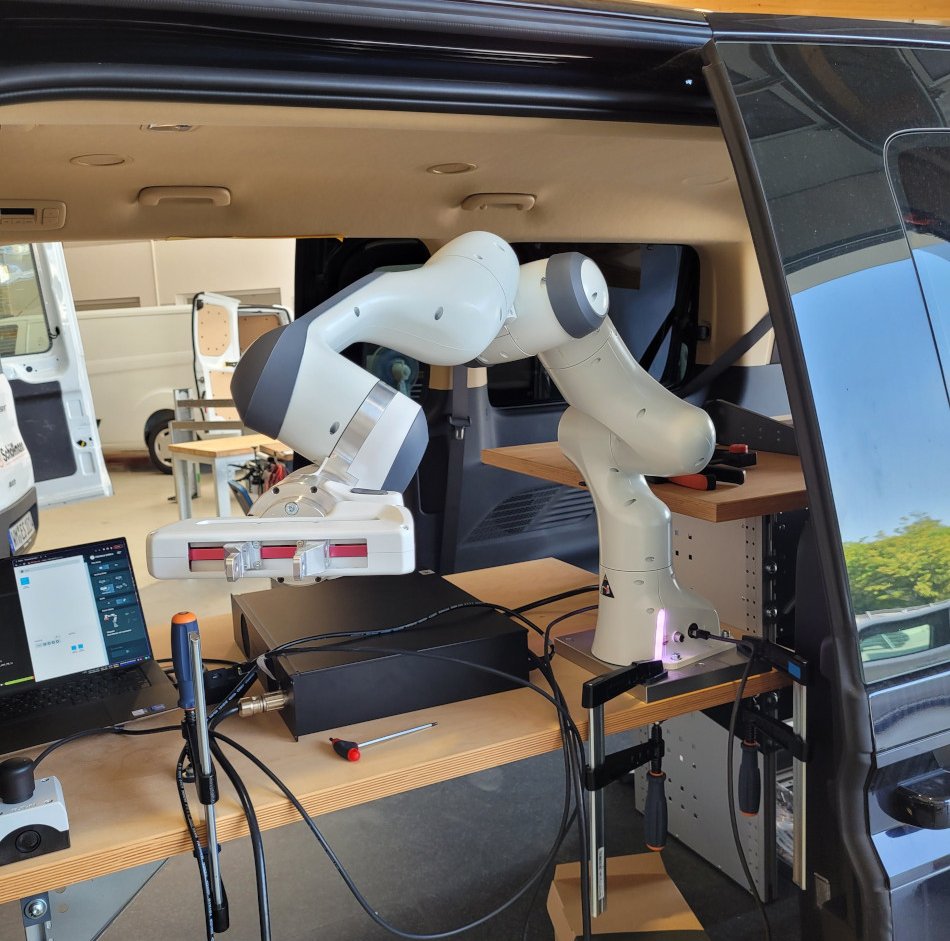}
    \caption[]{Lesson learned: Perform a robot fitting to ensure the robot has enough clearance from the ceiling.}
    \label{fig:svan-robot-fitting}
\end{figure}

\begin{itemize}
\item Ensuring the safe integration of additional interior features into a passenger vehicle requires careful planning and expertise, given the inherent complexity.
\item The retrofitting process for converting a standard passenger van involves significant challenges, notably achieving a secure installation of new flooring and fixtures.
\item It is advisable to opt for a commercial vehicle or engage the services of a professional specializing in vehicle equipment installation to ensure optimal safety and functionality.
\item Adequate vertical space must be allocated within the vehicle design to accommodate the installation and operational requirements of tactile robots (Fig. \ref{fig:svan-robot-fitting}).
\item Incorporate an ample number of power outlets within the design to support the electrical needs of all onboard equipment and devices.
\item Utilizing a magnetic mount for the antenna provides a straightforward and effective solution for its placement on the vehicle roof, ensuring stability and ease of installation.
\end{itemize}

\subsection{Field Trials}\label{sec:field_trials}

\begin{figure}
    \centering
    \includegraphics[width=\columnwidth, clip]{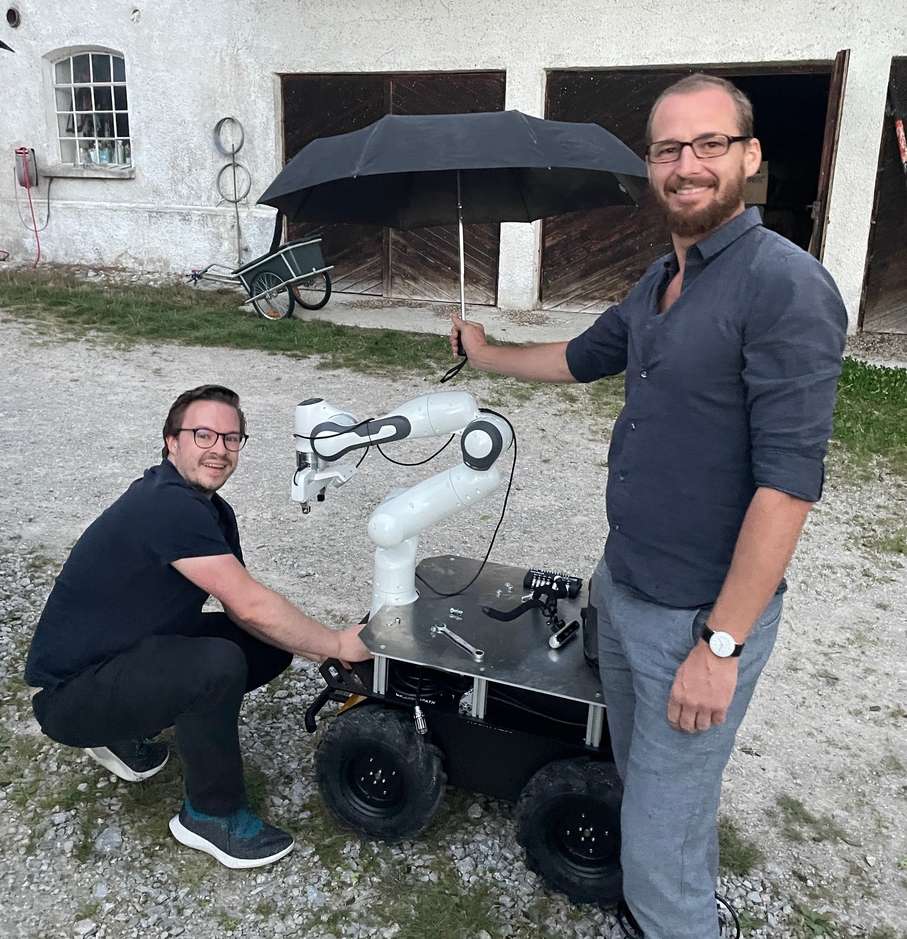}
    \caption[]{Lesson learned: An umbrella can safe a lot of money if used correctly.}
    \label{fig:panda-umbrella}
\end{figure}


\begin{itemize}
\item Anticipate software glitches; a larger battery capacity within the hub significantly enhances operational resilience and flexibility.
\item Implementing passthrough battery charging for the hub is crucial, allowing for power supply connectivity without necessitating a system shutdown, thereby ensuring continuous operation.
\item Tactile robotic manipulators are prone to damage during transport, and their mounting and unmounting processes can be laborious and time-consuming.
\item Maintaining backup computers (with fully installed environment) on hand offers a strategic advantage, ensuring operational continuity in the face of hardware failures.
\item While ROS 2 offers considerable benefits for robotic operations, it's not without its challenges, especially those that may arise unexpectedly in field conditions. Having the capability to address these issues on-site is invaluable.
\item A streamlined and well-organized networking setup is essential for efficient troubleshooting, allowing for quicker identification and resolution of connectivity issues.
\item Always be prepared for inclement weather when operating non-waterproof robots; packing an umbrella can provide necessary protection for your equipment (as depicted in Fig. \ref{fig:panda-umbrella}).
\end{itemize}

\section{Conclusion}\label{conclusion}
%

The journey of transforming a passenger vehicle into a functional mobile robot hub, as documented in this paper, unequivocally demonstrates its feasibility. However, this endeavor is not without its challenges and caveats, which we have meticulously identified and addressed throughout our research. While the prospect of fully autonomous deployment holds promise, it is contingent upon the advancements in autonomous driving technologies. Until such a time, our focus remains on minimizing the deployment time of robots, streamlining operations to enhance efficiency and effectiveness. Additionally, our experiences underline the need for an improved transport solution for tactile robots, ensuring their safety and functionality while mounted. Looking forward, the insights and lessons learned from developing and implementing SVAN pave the way for future innovations in environmental robotics, offering a glimpse into the potential transformations within the field.


%
%
\bibliographystyle{IEEEtran}
\bibliography{my_bib}

\end{document}